# Dynamic QoS Prediction via a Non-Negative Tensor Snowflake Factorization

YongHui Xia, Lan Wang*, Hao Wu

## Abstract

Dynamic quality of service (QoS) data exhibit rich temporal patterns in user-service interactions, which are crucial for a comprehensive understanding of user behavior and service conditions in Web service. As the number of users and services increases, there is a large amount of unobserved QoS data, which significantly affects users' choice of services. To predict unobserved QoS data, we propose a Non-negative Snowflake Factorization of tensors model. This method designs a snowflake core tensor to enhance the model's learning capability. Additionally, it employs a single latent factor-based, nonnegative multiplication update on tensor (SLF-NMUT) for parameter learning. Empirical results demonstrate that the proposed model more accurately learns dynamic user-service interaction patterns, thereby yielding improved predictions for missing QoS data.

**Keywords:** Quality of service (QoS) prediction, Tensor decomposition, non-negative constraint

## 1 Introduction

Cloud computing [1,2] and other service-oriented technological innovations have spurred an exponential increase in web services. However, as service functionalities become increasingly homogeneous, guaranteeing high-quality service (QoS) for end users remains a significant challenge [3]. QoS can be evaluated from the user's perspective (e.g., response time and throughput). Consequently, analyzing historical QoS data is critical for selecting the optimal service for a given task. In recent years, numerous methods that leverage QoS metrics for service selection have been proposed [4-7].

The rising costs and overwhelming number of services have rendered comprehensive evaluations in real-world environments both prohibitively expensive and time intensive. As a result, predicting missing QoS data to support evaluations has become a more feasible strategy. In this context, various matrix factorization [41-45] techniques for QoS prediction have been developed [8-11]. These approaches typically extract latent representations for users and services from sparse QoS matrices assumed to reside in a low-dimensional space, and then predict missing values by computing the inner products of the derived latent features [4,6].

It is important to note that the user-service QoS matrix is inherently sparse, since no user interacts with every available service and no service is accessed by all users. This inherent sparsity poses considerable challenges for achieving accurate and efficient representation learning [12]. To address this issue, researchers have introduced density-driven learning methods that have proven both effective and efficient [4,6]. For example, Luo et al. [13] presented a robust QoS prediction framework that utilizes an ensemble of diverse nonnegative latent factor analysis predictors, further accelerating the training process through sequential task learning [15] combined with alternating direction method solutions [14].

In particular, QoS experienced by a user for a given service is not static but varies over time. Many existing latent factor analysis models build fixed representations that fail to incorporate these temporal dynamics [16,20,21]. Extending these methods into a tensor factorization framework that integrates user, service, and time dimensions has shown promise. Hence, this study proposes a Non-negative Snowflake Factorization of Tensors (NSFT) model for dynamic QoS Prediction that captures the intricate dependencies in long-term, incomplete three-dimensional QoS datasets [22-26]. In our framework, the snowflake core tensor is leveraged to characterize the interactions among users, services, and time. Furthermore, we employ a single latent factor-based, nonnegative multiplication update on tensor (SLF-NMUT) for parameters learning [61-65]. The key contributions of this paper are as follows:

1) We present a NSFT model. It is able to accurately predict unobserved QoS data, where a snowflake core tensor is introduced to model the relationship of latent features.

2) We provide a detailed algorithmic design and an extensive empirical evaluation of the proposed model.

## 2 Preliminaries

### 2.1 QoS tensor

Given three positive integer sets $I = \{i : i \in \mathbb{Z}^+\}$, $J = \{j: j \in \mathbb{Z}^+\}$ and $K = \{k : k \in \mathbb{Z}^+\}$, $\mathbf{Y}^{|I| \times |J| \times |K|}$ is a user-service-time QoS tensor and each element $y_{ijk} \in \mathbb{R}_0^+$ denotes the QoS experienced by a user $i$ invoking a service $j$ during a time slice $k$.

Due to the incomplete and non-negative nature of QoS data, such a tensor is only partially filled with non-negative real numbers, as illustrated in Fig. 1. Note that, in this paper, unless otherwise specified, the time dimension refers to a discrete set of time slices [46-50].

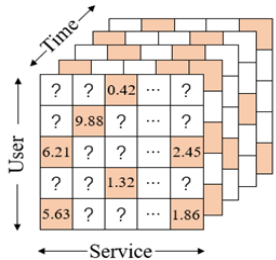 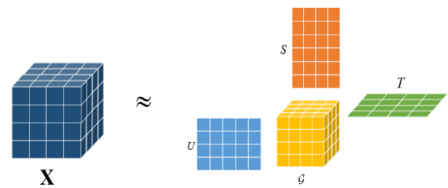

Fig. 1.　QoS tensor of user–service–time showing the response time.　　　　Fig. 2. The schematic diagram of the Tucker decomposition model.

## 2.2 Tucker decomposition

A tensor can be viewed as an array extending across multiple dimensions, and the count of these dimensions (often referred to as modes or rank) characterizes its structure [56-60]. Let $\mathbf{X} \in \mathbb{R}^{I \times J \times K}$ represent an input tensor, where $I$ and $J$ indicate entity-related dimensions, while $K$ corresponds to the temporal dimension [51-55]. Suppose we have a High-Dimension and Incomplete (HDI) tensor $\mathbf{Y}$ [34-40]. We denote by $\Lambda$ the collection of observed entries and by $\Gamma$ the (larger) collection of unobserved entries, so $|\Gamma| \gg |\Lambda|$. Tucker decomposition is noteworthy for encompassing all feature interactions and for generating a core tensor that supplies weights to each rank-one component. The schematic diagram of the Tucker decomposition model is shown in Fig. 2. In this paper, we express the approximation as

$$\hat{\mathbf{Y}} = \sum_{p=1}^{P} \sum_{q=1}^{Q} \sum_{r=1}^{R} g_{pqr} \mathcal{A}_{pqr}, \tag{1}$$

where $g_{pqr}$ is the $(p,q,r)$-th entry of the core tensor $\mathcal{G}$, indicating the magnitude of each feature interaction, and $\mathcal{A}_{pqr}$ is formed by combining the latent vectors $a_p \in \mathbb{R}^I$, $b_q \in \mathbb{R}^J$, and $c_r \in \mathbb{R}^K$ for their respective dimensions [17].

## 3 Non-negative Snowflake Factorization of Tensors Model

### 3.1 Model Design

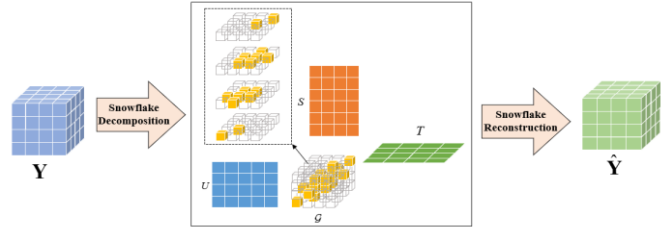

Fig. 3. The schematic design of NSFT model.

This NSFT model leverages the snowflake core tensor to enhance the model's learning capability. The schematic design of NSFT model is shown in Fig. 3. The element-wise representation of the NSFT is expressed as follows:

$$\hat{y}_{ijk} = \sum_{r=1}^{R} \Big( g_{rrr} u_{ir} s_{jr} t_{kr} + \sum_{f=1}^{F} \Big( g_{(r-f)rr} u_{i(r-f)} s_{jr} t_{kr} + g_{r(r-f)r} u_{ir} s_{j(r-f)} t_{kr} + g_{rr(r-f)} u_{ir} s_{jr} t_{k(r-f)} \\ + g_{(r+f)rr} u_{i(r+f)} s_{jr} t_{kr} + g_{r(r+f)r} u_{ir} s_{j(r+f)} t_{kr} + g_{rr(r+f)} u_{ir} s_{jr} t_{k(r+f)} \Big) \Big). \tag{2}$$

An objective function is set to find an optimal set of LF matrices $U$, $S$ and $T$, by measuring the difference between $\mathbf{Y}$ and $\hat{\mathbf{Y}}$ with Euclidean distance in this paper. It is given by

$$\varepsilon = \left\| \mathbf{Y} - \hat{\mathbf{Y}} \right\|_F^2, \tag{3}$$

where $\varepsilon$ denotes the objective function and $\|\cdot\|$ is the Frobenius norm operator.

Inherently, (3) is ill-posed because the known elements of **Y** are distributed unevenly, making the solution highly sensitive to the initial assumptions of snowflake core tensor and $U$, $S$, $T$. To enhance the model's robustness, Tikhonov regularization is introduced, and the objective is subsequently reformulated in a revised manner:

$$\varepsilon = \left(y_{ijk} - \hat{y}_{ijk}\right)^2 + H,$$

$$\hat{y}_{ijk} = \sum_{r=1}^{R}\Big(g_{rrr}u_{ir}s_{jr}t_{kr} + \sum_{f=1}^{F}\big(g_{(r-f)rr}u_{i(r-f)}s_{jr}t_{kr} + g_{r(r-f)r}u_{ir}s_{j(r-f)}t_{kr} + g_{rr(r-f)}u_{ir}s_{jr}t_{k(r-f)}$$
$$+ g_{(r+f)rr}u_{i(r+f)}s_{jr}t_{kr} + g_{r(r+f)r}u_{ir}s_{j(r+f)}t_{kr} + g_{rr(r+f)}u_{ir}s_{jr}t_{k(r+f)}\big)\Big),$$
(4)

$$H = \lambda_a \sum_{r=1}^{R}\Big(g_{rrr}^2 + \sum_{f=1}^{F}\big(g_{(r-f)rr}^2 + g_{r(r-f)r}^2 + g_{rr(r-f)}^2 + g_{(r+f)rr}^2 + g_{r(r+f)r}^2 + g_{r(r+f)r}^2\big)\Big) + \lambda_b \sum_{r=1}^{R}\big(u_{ir}^2 + s_{jr}^2 + t_{kr}^2\big).$$

It is evident that QoS data exhibit temporal fluctuations, with each user-service pair displaying its own distinctive pattern of change over time. Consequently, to effectively examine such variability, it becomes crucial to incorporate linear bias (LB) terms into the objective function for enhancing both the stability and expressive power of a learning framework [32].

Three matrices $\mathcal{Q}_u^{|I|\times 3}$, $\mathcal{Q}_s^{|J|\times 3}$, and $\mathcal{Q}_t^{|K|\times 3}$ serve as the foundation for constructing the LBs. Each matrix's columns encode distinct bias structures for user, service, or time [17-19]. Within every LB matrix, a single "active" column contains tunable bias parameters for the training process, whereas the other two columns remain fixed at 1. By applying the outer product to these LB column vectors along a given dimension, three rank-one bias tensors $\mathcal{Q}_u$, $\mathcal{Q}_s$, and $\mathcal{Q}_t$ are generated and incorporated into the low-rank approximation of **Y**, and the element-wise form of $\hat{\mathbf{Y}}$ is formulated as

$$\hat{y}_{ijk} = \sum_{r=1}^{R}\Big(g_{rrr}u_{ir}s_{jr}t_{kr} + \sum_{f=1}^{F}\big(g_{(r-f)rr}u_{i(r-f)}s_{jr}t_{kr} + g_{r(r-f)r}u_{ir}s_{j(r-f)}t_{kr} + g_{rr(r-f)}u_{ir}s_{jr}t_{k(r-f)}$$
$$+ g_{(r+f)rr}u_{i(r+f)}s_{jr}t_{kr} + g_{r(r+f)r}u_{ir}s_{j(r+f)}t_{kr} + g_{rr(r+f)}u_{ir}s_{jr}t_{k(r+f)}\big)\Big) + q_{ijk}^u + q_{ijk}^s + q_{ijk}^t.$$
(6)

Because $\mathcal{Q}_u$, $\mathcal{Q}_s$, and $\mathcal{Q}_t$ each have only a single active dimension, it becomes more suitable to reshape them into three corresponding active bias vectors $a^{|I|}$, $b^{|J|}$, and $c^{|K|}$ in accordance with the mapping described below:

$$q_{ijk}^u = a_i, \forall i \in I; \; q_{ijk}^s = b_j, \forall j \in J; \; q_{ijk}^t = c_k, \forall k \in K.$$
(7)

Thus, (6) is simplified into

$$\hat{y}_{ijk} = \sum_{r=1}^{R}\Big(g_{rrr}u_{ir}s_{jr}t_{kr} + \sum_{f=1}^{F}\big(g_{(r-f)rr}u_{i(r-f)}s_{jr}t_{kr} + g_{r(r-f)r}u_{ir}s_{j(r-f)}t_{kr} + g_{rr(r-f)}u_{ir}s_{jr}t_{k(r-f)}$$
$$+ g_{(r+f)rr}u_{i(r+f)}s_{jr}t_{kr} + g_{r(r+f)r}u_{ir}s_{j(r+f)}t_{kr} + g_{rr(r+f)}u_{ir}s_{jr}t_{k(r+f)}\big)\Big) + a_i + b_j + c_k.$$
(8)

By inserting (8) into (5) and incorporating regularization for the bias elements, one arrives at a biased variant of the objective function as follows:

$$\varepsilon = \left(y_{ijk} - \hat{y}_{ijk}\right)^2 + H,$$

$$\hat{y}_{ijk} = \sum_{r=1}^{R}\Big(g_{rrr}u_{ir}s_{jr}t_{kr} + \sum_{f=1}^{F}\big(g_{(r-f)rr}u_{i(r-f)}s_{jr}t_{kr} + g_{r(r-f)r}u_{ir}s_{j(r-f)}t_{kr} + g_{rr(r-f)}u_{ir}s_{jr}t_{k(r-f)}$$
$$+ g_{(r+f)rr}u_{i(r+f)}s_{jr}t_{kr} + g_{r(r+f)r}u_{ir}s_{j(r+f)}t_{kr} + g_{rr(r+f)}u_{ir}s_{jr}t_{k(r+f)}\big)\Big) + a_i + b_j + c_k,$$

$$H = \lambda_a \sum_{r=1}^{R}\Big(g_{rrr}^2 + \sum_{f=1}^{F}\big(g_{(r-f)rr}^2 + g_{r(r-f)r}^2 + g_{rr(r-f)}^2 + g_{(r+f)rr}^2 + g_{r(r+f)r}^2 + g_{rr(r+f)}^2\big)\Big) + \lambda_b \sum_{r=1}^{R}\big(u_{ir}^2 + s_{jr}^2 + t_{kr}^2\big) + \lambda_c\big(a_i^2 + b_j^2 + c_k^2\big),$$

$$s.t. \forall i \in I, j \in J, k \in K, r \in \{1,2,...,R\}: f < r, f < R-r, g_{rrr} \geq 0, g_{(r-f)rr} \geq 0, g_{r(r-f)r} \geq 0,$$
$$g_{rr(r-f)} \geq 0, g_{(r+f)rr} \geq 0, g_{r(r+f)r} \geq 0, g_{rr(r+f)} \geq 0, u_{ir} \geq 0, s_{jr} \geq 0, t_{kr} \geq 0, a_i \geq 0, b_j \geq 0, c_k \geq 0.$$

(9)

### 3.2 Parameter Learning via SLF-NMUT

Due to the non-negative nature of QoS data, the model's parameters must also remain non-negative. Previous investigations demonstrate that SLF-NMUT is both effective and efficient for realizing non-negative LFA on HDI data. Nevertheless, integrating SLF-NMUT into our snowflake decomposition model requires a thorough examination of the relevant parameter learning rules [4,28].

To begin, gradient descent is performed on (9), producing learning rules for every parameter in the snowflake core tensor as well as in the LF and LB matrices [27-31]. Because the update patterns for each LF matrix are essentially alike and the same applies to the LB vectors—we present only the update formulas for a specified element in the snowflake core tensor and $U$, $a$:

$$\begin{cases} g_{rrr} \leftarrow g_{rrr} - \eta_g \dfrac{\partial \varepsilon}{\partial g_{rrr}} = g_{rrr} - \eta_g\left(-\theta u_{ir}s_{jr}t_{kr} + \lambda_g g_{rrr}\right), \\ u_{ir} \leftarrow u_{ir} - \eta_{ir}\left(-\theta\Big(g_{rrr}s_{jr}t_{kr} + \sum_{f=1}^{F}\big(g_{r(r-f)r}s_{j(r-f)}t_{kr} + g_{rr(r-f)}s_{jr}t_{k(r-f)} + g_{r(r+f)r}s_{j(r+f)}t_{kr} + g_{rr(r+f)}s_{jr}t_{k(r+f)}\big)\Big) + \lambda_b u_{ir}\right), \\ a_i \leftarrow a_i - \eta_i\left(-\theta + \lambda_c a_i\right). \end{cases}$$

(10)

where $\theta = y_{ijk} - \hat{y}_{ijk}$, $\eta_g$, $\eta_{ir}$ and $\eta_i$ are learning rates for $g_{rrr}$, $u_{ir}$ and $a_i$ respectively.

In accordance with SLF-NMUT guidelines, the learning rates are tuned to eliminate their respective negative components. As a concrete illustration of our final configuration, the values of $\eta_g$ and $\eta_i$ are set as follows:

$$\begin{cases} \eta_g = \dfrac{g_{rrr}}{\hat{y}_{ijk}u_{ir}s_{jr}t_{kr} + \lambda_a g_{rrr}}, \\ \eta_i = \dfrac{a_i}{\hat{y}_{ijk} + \lambda_c a_i}. \end{cases}$$

(11)

Following these modifications, a group of non-negative multiplicative update formulas emerges for the model parameters. Specifically, the update equations for $g_{rrr}$, $u_{ir}$ and $a_i$ are as follows:

$$\begin{cases} g_{rrr} \leftarrow \dfrac{g_{rrr} y_{ijk} u_{ir} s_{jr} t_{kr}}{\hat{y}_{ijk} u_{ir} s_{jr} t_{kr} + \lambda_a g_{rrr}}, \\ u_{ir} \leftarrow \dfrac{u_{ir} y_{ijk} \left(g_{rrr} s_{jr} t_{kr} + \beta\right)}{\hat{y}_{ijk} \left(g_{rrr} s_{jr} t_{kr} + \beta\right) + \lambda_b u_{ir}}, \\ a_i \leftarrow \dfrac{a_i y_{ijk}}{\hat{y}_{ijk} + \lambda_c a_i}. \end{cases} \quad (12)$$

where $\beta = \sum_{f=1}^{F} \left(g_{r(r+f)r} s_{j(r+f)} t_{kr} + g_{rr(r+f)} s_{jr} t_{k(r+f)} + g_{r(r-f)r} s_{j(r-f)} t_{kr} + g_{rr(r-f)} s_{jr} t_{k(r-f)}\right).$

## 4 Experiments

### 4.1 Datasets

In order to assess how effectively our Snowflake Decomposition model performs, we rely on two authentic QoS datasets gathered by WS-DREAM [35]. Table I summarizes these datasets, which document the response time and throughput of 4500 real-world web services utilized by 142 users across 64 consecutive time slices. Each dataset contains 30,287,611 service invocation logs and maintains a uniform density of 74.06%, a figure comparatively high for real-world scenarios. Table II provides further details, QoS entries equating to zero are excluded from the assessment.

TABLE I. DETAILS OF WSDREAM DATASETS

| Dataset | Qos-RT | Qos-TP |
|---|---|---|
| Data Type | Response time | Throughput |
| Value Scale | 0-20 s | 0-1000 kbps |
| User Count | 142 | 142 |
| Service Count | 4500 | 4500 |
| Time slice Count | 64 | 64 |
| Record Count | 30,287,611 | 30,287,611 |

TABLE II. DETAILS OF EXPERIMENT DATASETS

| Dataset | No. | Train:Valid:Test |
|---|---|---|
| Qos-RT | D1 | 1:2:7 |
|  | D2 | 2:2:6 |
| Qos-TP | D3 | 1:2:7 |
|  | D4 | 2:2:6 |

### 4.2 Evaluation Metrics

Evaluating QoS prediction accuracy is essential, as it demonstrates how effectively the model captures key QoS features within user-service-time interactions. Accordingly, this study emphasizes model accuracy and employs mean absolute error (MAE) along with root mean square error (RMSE) for quantitative evaluation:

$$MAE = \dfrac{\sum_{y_{ijk} \in \Lambda} \left|y_{ijk} - \hat{y}_{ijk}\right|}{|\Lambda|}, \quad RMSE = \sqrt{\sum_{y_{ijk} \in \Lambda} \dfrac{\left(y_{ijk} - \hat{y}_{ijk}\right)^2}{|\Lambda|}}. \quad (13)$$

Throughout training, metrics on the validation set are computed to assess whether the model has converged. In this study, convergence is assumed once the difference in validation errors between successive iterations drops below. Subsequently, metrics are computed on the test set and employed for performance comparisons.

### 4.3 Model Comparison

The performance of our proposed Snowflake Decomposition model is evaluated in comparison with several cutting-edge models that are capable of providing accurate predictions for missing data in high-dimensional and incomplete tensors. The models used for comparison include: (a) M1: CP-SGD [34]; (b) M2: NNCP [28]; (c) M3: CTF [5]; (d) M4: Our NSFT model.

TABLE III. THE SUMMARY OF RESULTS

|    | D1 |        | D2 |        | D3 |         | D4 |         |
|----|--------|--------|--------|--------|--------|---------|--------|---------|
|    | MAE    | RMSE   | MAE    | RMSE   | MAE    | RMSE    | MAE    | RMSE    |
| M1 | 1.4746 | 3.1152 | 1.4863 | 3.0752 | 4.5475 | 25.4634 | 4.5579 | 24.7702 |
| M2 | 1.4347 | 3.0835 | 1.4346 | 3.0848 | 4.4621 | 25.3390 | 4.3927 | 24.5312 |
| M3 | 1.4331 | 3.0814 | 1.4308 | 3.0789 | 4.4652 | 24.9702 | 4.3961 | 24.4137 |
| M4 | 1.4315 | 3.0768 | 1.4293 | 3.0704 | 4.3979 | 24.4542 | 4.3775 | 24.0865 |

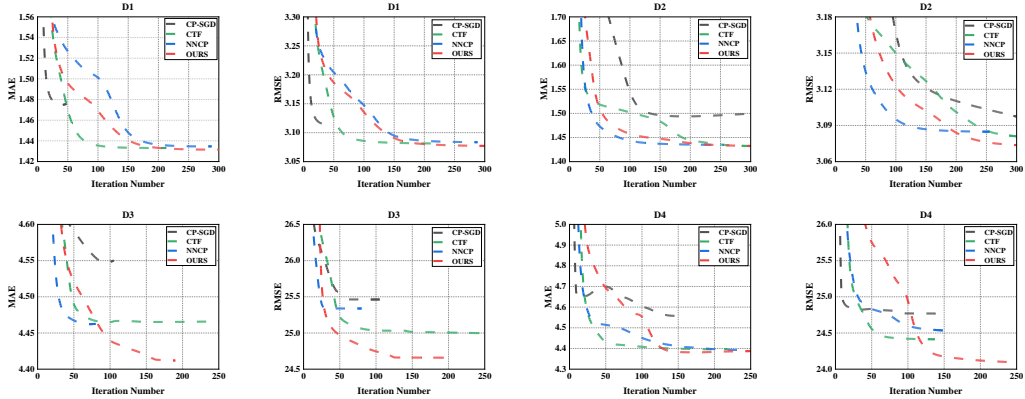

Fig. 4. The MAE and RMSE of M1, M2, M3 and M4 on D1, D2, D3 and D4.

In this experiment, all the results are summarized in Table III and Fig. 4. Based on these results, the following main conclusions can be drawn:

1) Our NSFT model performs exceptionally well in handling high-dimensional and incomplete data. As shown in Table III, For instance, on the D1 dataset, M4 achieves improvements in MAE of 2.9%, 0.2%, and 0.1% over M1, M2, and M3, respectively, M4 achieves improvements in RMSE of 1.2%, 0.2%, and 0.1% over M1, M2, and M3, respectively. On the D2 dataset, M4 achieves improvements in MAE of 3.8%, 0.3%, and 0.1% over M1, M2, and M3, respectively, and improvements in RMSE of 0.1%, 0.4%, and 0.2% over M1, M2, and M3, respectively. On the D3 dataset, M4 outperforms M1, M2, and M3 by achieving MAE improvements of 3.2%, 1.4%, and 1.5%, respectively, and RMSE enhancements of 3.9%, 3.4%, and 2% over M1, M2, and M3, respectively. On the D4 dataset, M4 demonstrates MAE improvements of 3.9%, 0.3%, and 0.4% over

M1, M2, and M3, respectively, and it achieves RMSE enhancements of 2.7%, 1.8%, and 1.3% compared to M1, M2, and M3, respectively. This implies that the NSFT model is also a valid low-rank tensor completion model.

2) From the experimental results and analysis, it can be seen that with testing on various types of data, the NSFT model offers better interpretability and generalization capabilities at the spatiotemporal or representational level. Moreover, its prediction accuracy in scenarios with high-dimensional and missing data outperforms traditional models (such as M1–M3), further demonstrating its applicability across various tasks.

3) The NSFT model demonstrates robust performance under varying data sparsity levels and parameter settings, highlighting its adaptability in diverse real-world scenarios. Even when the proportion of missing entries increases, the model maintains high predictive accuracy, suggesting its practical viability for large-scale QoS datasets.

# 5  Conclusion

In this paper, we introduced a novel tensor decomposition method termed the Non-negative Snowflake Factorization of Tensors model for dynamic QoS prediction. Our model leverages a unique snowflake core tensor to capture the complex interdependencies among users, services, and time, thereby accurately modeling the intrinsic spatiotemporal variations in QoS data. To further enhance the stability and predictive performance under nonnegative constraints, we adopted a single latent factor-based nonnegative tensor multiplicative update (SLF-NMUT) strategy for parameter learning.

Extensive experiments conducted on real-world QoS datasets demonstrate that our NSFT model consistently outperforms conventional low-rank tensor completion methods in terms of both MAE and RMSE. Notably, the model exhibits robust performance even under conditions of high dimensionality, significant data incompleteness, and varying levels of sparsity. This indicates its strong generalization capabilities and high interpretability when applied to complex, dynamic service environments. Our work also includes a detailed presentation of the algorithmic design and parameter learning rules, which not only reduce computational complexity and training costs but also maintain high predictive accuracy. The experimental findings confirm that the model is capable of handling substantial amounts of missing data, making it a practical solution for large-scale QoS prediction tasks.

In future work, we plan to extend the applicability of the proposed framework to a wider range of real-world scenarios and larger datasets. Future work will explore the integration of advanced optimization strategies and deep learning techniques to further improve the model's ability to capture intricate temporal dynamics. Moreover, we aim to incorporate external information, such as user behavior data and environmental variables, to enrich the model's representational power and achieve even more precise predictions.